\documentclass{article}

\usepackage[preprint]{neurips_2019}

\usepackage[utf8]{inputenc} % allow utf-8 input
\usepackage[T1]{fontenc}    % use 8-bit T1 fonts
\usepackage{hyperref}       % hyperlinks
\usepackage{url}            % simple URL typesetting
\usepackage{booktabs}       % professional-quality tables
\usepackage{amsfonts}       % blackboard math symbols
\usepackage{nicefrac}       % compact symbols for 1/2, etc.
\usepackage{microtype}      % microtypography
\usepackage{graphicx,wrapfig}
\usepackage[skip=2pt,font=footnotesize,labelfont=bf]{caption}
\usepackage{subcaption}
\usepackage{amsmath,amssymb}
\usepackage{array}
\usepackage{wrapfig}

\usepackage[capitalize]{cleveref}
\crefname{section}{Sec.}{Secs.}
\Crefname{section}{Section}{Sections}
\Crefname{table}{Table}{Tables}
\crefname{table}{Tab.}{Tabs.}

\newcolumntype{U}{>{\centering\arraybackslash}p{2.90cm}}
\newcolumntype{V}{>{\centering\arraybackslash}p{2.00cm}}
\newcolumntype{T}{>{\centering\arraybackslash}p{1.0cm}}

\title{Temporal Embeddings: Scalable Self-Supervised Temporal Representation Learning from Spatiotemporal Data for Multimodal Computer Vision}

\author{%
    Yi Cao\thanks{Equal Contribution}, Swetava Ganguli\footnotemark[1], Vipul Pandey \\
    Apple\\
    \texttt{\{ycao4,swetava,vipul\}@apple.com} \\
}

\begin{document}

%%%%%%%%%%%%%%%%%%%%%%%%%%%%%%%%%%%%%%%%%%%%%%%%%%%%%%%%%%%%%%%%%%%%%%%%
% Make the title
%%%%%%%%%%%%%%%%%%%%%%%%%%%%%%%%%%%%%%%%%%%%%%%%%%%%%%%%%%%%%%%%%%%%%%%%
% Make the title
\maketitle

%%%%%%%%%%%%%%%%%%%%%%%%%%%%%%%%%%%%%%%%%%%%%%%%%%%%%%%%%%%%%%%%%%%%%%%%
% Section: Introduction
%%%%%%%%%%%%%%%%%%%%%%%%%%%%%%%%%%%%%%%%%%%%%%%%%%%%%%%%%%%%%%%%%%%%%%%%
\setlength{\belowcaptionskip}{0pt}
\setlength{\textfloatsep}{0pt}
\begin{wrapfigure}{L}{0.30\textwidth}
    \centering
    \includegraphics[width=0.30\textwidth]{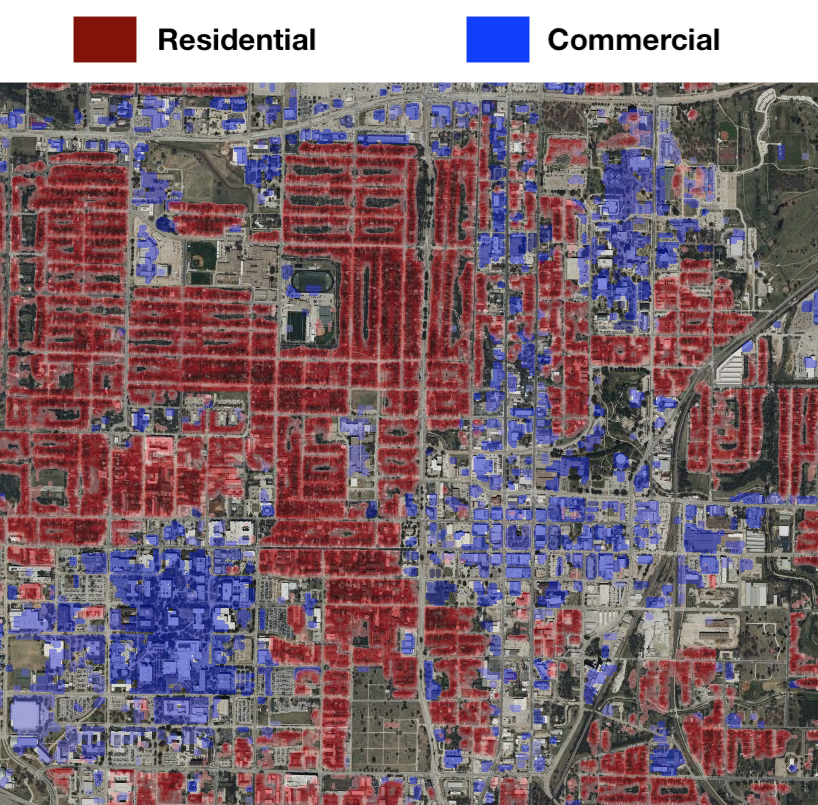}
    \caption{Segmentation of residential and commercial areas using temporal embeddings.}
    \label{figure::residential_commercial}
\end{wrapfigure}

\vspace{-1.0cm}
\section{Introduction}\label{section::introduction}
\vspace{-0.3cm}
There exists a correlation between geospatial activity temporal patterns and type of land use. Conducting proper data mining on mobility data to survey geographic landscape has been proven to be as cost-efficient as other traditional remote sensing techniques. By analyzing user activity volume change over time at different locations, we found that commercial area exhibits various heartbeat patterns while residential area largely shows randomness. This observation reveals a correlation between activity temporal patterns and type of land use, which motivates us to further use a self-supervised framework to extract temporal features from mobility data and apply the features to landscape classification. A novel self-supervised approach is proposed to stratify landscape based on mobility activity time series. First, the time series signal is transformed to the frequency domain and then compressed into task-agnostic \textit{temporal embeddings} by a contractive autoencoder, which preserves cyclic temporal patterns observed in time series. The pixel-wise embeddings are converted to image-like channels that can be used for task-based, multimodal modeling of downstream geospatial tasks using deep semantic segmentation on a geospatial AI platform such as \cite{iyer2021trinity,iyer2023perspectives}. Experiments show that temporal embeddings are semantically meaningful representations of time series data and are effective across different tasks such as classifying residential and commercial areas, classifying activity areas like golf courses, grocery shops, road intersections, educational buildings, etc., and for stratifying the landscape into various activity stratas such as downtown, urban, suburban, rural, etc. Temporal embeddings transform sequential, spatiotemporal motion trajectory data into semantically meaningful image-like tensor representations that can be combined (multimodal fusion) with other data modalities (on machine learning platforms such as Trinity \cite{iyer2021trinity}) that are or can be transformed into image-like tensor representations (for e.g., RBG imagery, graph embeddings of road networks, passively collected imagery like SAR, etc.) to facilitate multimodal learning in geospatial computer vision. Multimodal computer vision is critical for training machine learning models for geospatial feature detection to keep a geospatial mapping service up-to-date in real-time and can significantly improve user experience and above all, user safety. At BayLearn, we wish to present a poster describing our work in \cite{cao2023self} that was presented at the 43rd IEEE International Geoscience and Remote Sensing Symposium (IEEE IGARSS) 2023.

% Tiles resulting from a raster representation of the earth's surface are modeled as nodes on a graph or pixels of an image. GPS trajectories are modeled as allowed Markovian paths on these nodes. A scalable and distributed algorithm is presented in \cite{ganguli2021reachability} to compute image-like tensor representations, called \textit{reachability summaries}, of the spatial connectivity patterns between tiles and their neighbors implied by the observed Markovian paths. A convolutional, contractive autoencoder is trained to learn compressed representations, called \textit{reachability embeddings}, of reachability summaries for every tile. 

%%%%%%%%%%%%%%%%%%%%%%%%%%%%%%%%%%%%%%%%%%%%%%%%%%%%%%%%%%%%%%%%%%%%%%%%
% Section: The Temporal Embeddings Algorithm Proposed in the Paper
%%%%%%%%%%%%%%%%%%%%%%%%%%%%%%%%%%%%%%%%%%%%%%%%%%%%%%%%%%%%%%%%%%%%%%%%
\section{The Temporal Embeddings Algorithm Proposed in \cite{cao2023self}}\label{section::temporalembeddings}
\begin{figure}
    \centering
    \includegraphics[width=\textwidth]{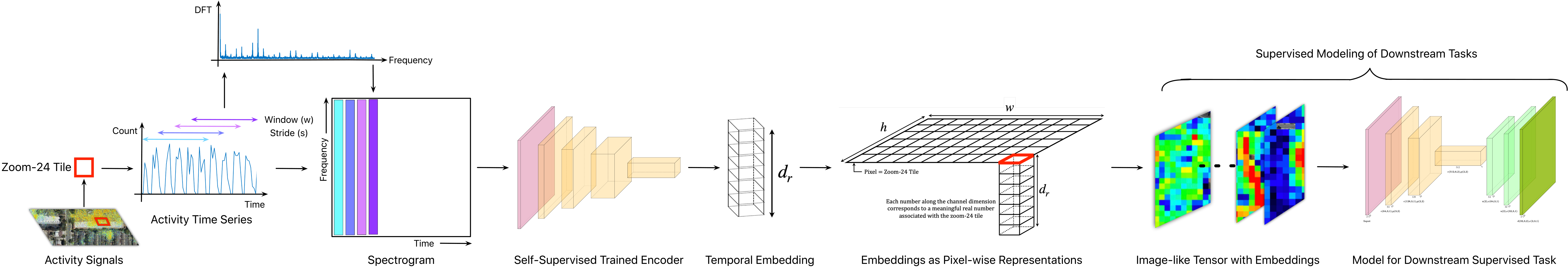}
    \caption{Schematic demonstrating the generation of temporal embeddings.}
    \label{figure::schematic}
\end{figure}

\vspace{-0.2cm}
A \textit{GPS trajectory} encodes spatiotemporal movement of an object as a chronologically ordered sequence of GPS records (a tuple of timestamp and the location's zoom-24 tile \cite{z24tiledefinition}). Let $\mathcal{T}$ represent the set of all available GPS trajectories during the time interval $[t_0, t_0+\Delta t]$ such that all GPS records in each trajectory are associated with the same motion modality (e.g., driving, walking, biking). The \textit{Earth Surface Graph} (ESG), $\boldsymbol{G}_{ES}(\boldsymbol{V}_{ES},\boldsymbol{E}_{ES})$, is defined as the inferred graph obtained from a raster representation of the spherical Mercator (WGS 84) projection \cite{epsg_3857} of earth's surface based on zoom-24 tiles \cite{z24tiledefinition} with these tiles as nodes, $\boldsymbol{V}_{ES}$. 
\setlength{\belowcaptionskip}{0pt}
\setlength{\textfloatsep}{0pt}
\begin{wrapfigure}{R}{0.30\textwidth}
    \centering
    \includegraphics[width=0.30\textwidth]{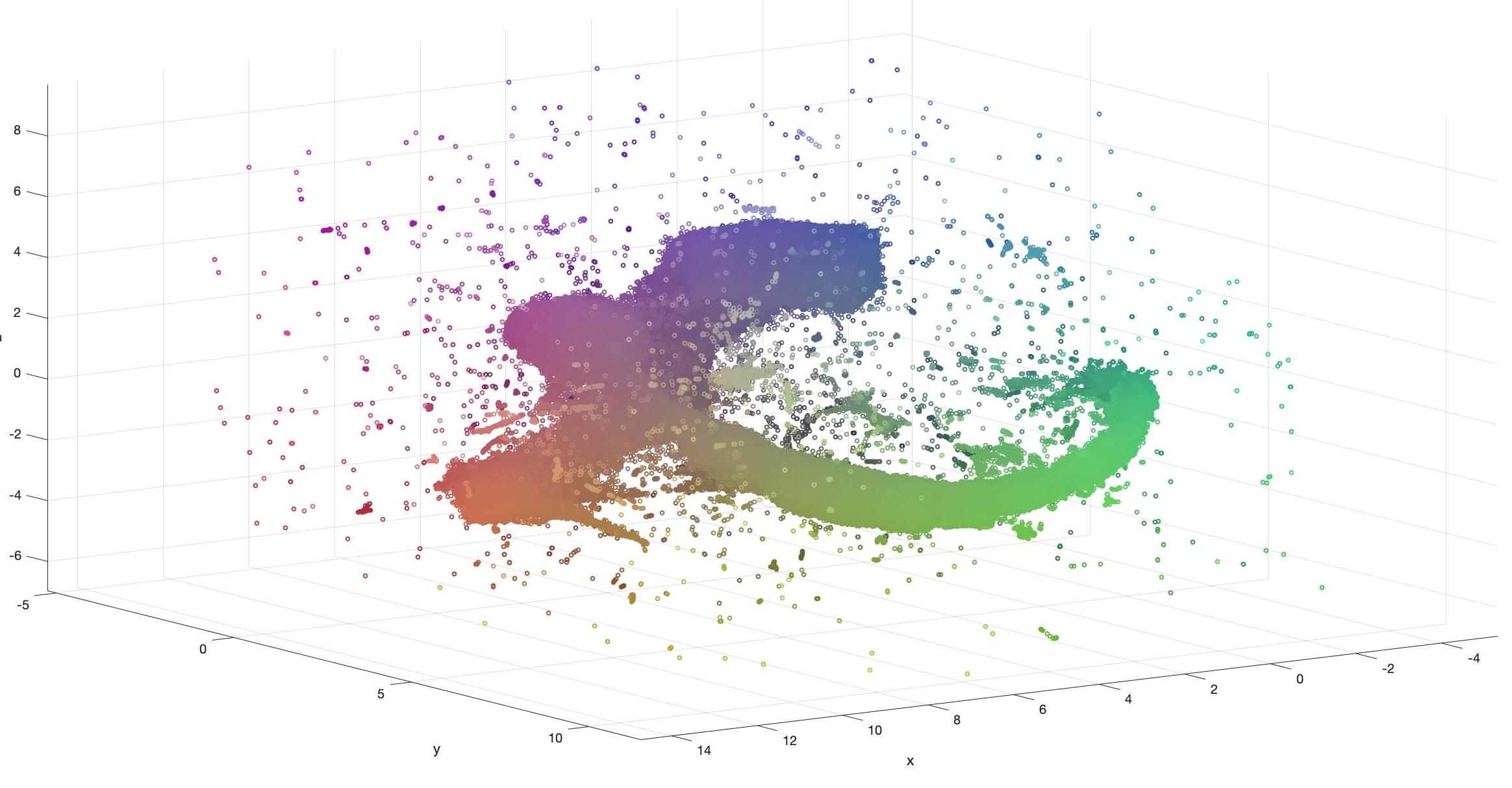}
    \caption{UMAP of learned temporal embeddings.}
    \label{umap}
\end{wrapfigure}
Using the count of GPS records per zoom-24 tile over a chosen period of time ($\Delta t$) for a chosen number of intervals ($\Delta t$, $2\Delta t$, $\ldots$, $N\Delta t$), we create time series profile for each zoom-24 tile and use Discrete Fourier Transform (DFT) to convert the activity signal from time domain into frequency domain. This feature engineering exposes temporal patterns that are characteristic to type of land use. But the resultant DFT vector could be too long ($\mathcal{O}(N)$) to be used as pixel-wise representations in convolutional neural networks (CNN) for downstream classification task (where each component of the representation becomes a channel in the input image-like tensor) when time series span a long time period with fine granularity. On the other hand, there are only a few frequency bands among the whole spectrum that are important (e.g., hourly, daily, weekly patterns) while the remainder are irrelevant noise (typically in high frequency spectrum). To create more compact and informative feature vectors, we adopt a rolling DFT window that captures short-term dynamics as well as long-term trends in time series and reshape the DFT array in a matrix like an image. We then employ a contractive autoencoder \cite{Goodfellow-et-al-2016} to compress the DFT image into a temporal embeddings vector, which preserve cyclic temporal patterns. Furthermore, given that the size of the vector embedding is controllable, we can choose a size that is a good compromise between storage constraints on feature stores and accuracy of downstream tasks. In other words, the input to the autoencoder is the spectrogram of the time series signal that captures the frequency response during a preceeding time window and the change in the frequency response as the time window is shifted. As a result, the learned compressed embeddings (16-dimensional) contain essential temporal features which are a more amenable input to segmentation neural networks \cite{iyer2021trinity} for classification task, as shown in \cref{figure::schematic}. As compared to using raw DFT as input, temporal embeddings significantly reduces the number of trainable parameters for the segmentation model, thereby making downstream models easier to train and converge. Other benefits attributed to the embeddings include: a) the embeddings encode spatiotemporal dynamics in a compact format that is easier for data storage in feature stores like \cite{iyer2021trinity}, b) temporal embeddings are task-agnostic and can be used to train different task-specific models for downstream tasks using geospatial AI platforms like \cite{iyer2021trinity} and other data modalities and/or learned representations \cite{xiao2020vae,ganguli2022reachability,reshetova2023semand}.  

%%%%%%%%%%%%%%%%%%%%%%%%%%%%%%%%%%%%%%%%%%%%%%%%%%%%%%%%%%%%%%%%%%%%%%%%
% Section: Qualitative and Quantitative Results
%%%%%%%%%%%%%%%%%%%%%%%%%%%%%%%%%%%%%%%%%%%%%%%%%%%%%%%%%%%%%%%%%%%%%%%%
\section{Qualitative and Quantitative Results}\label{section::results}
\begin{figure}[h]
    \centering
    \begin{minipage}[c]{0.24\textwidth}
        \centering
        \includegraphics[width=\textwidth]{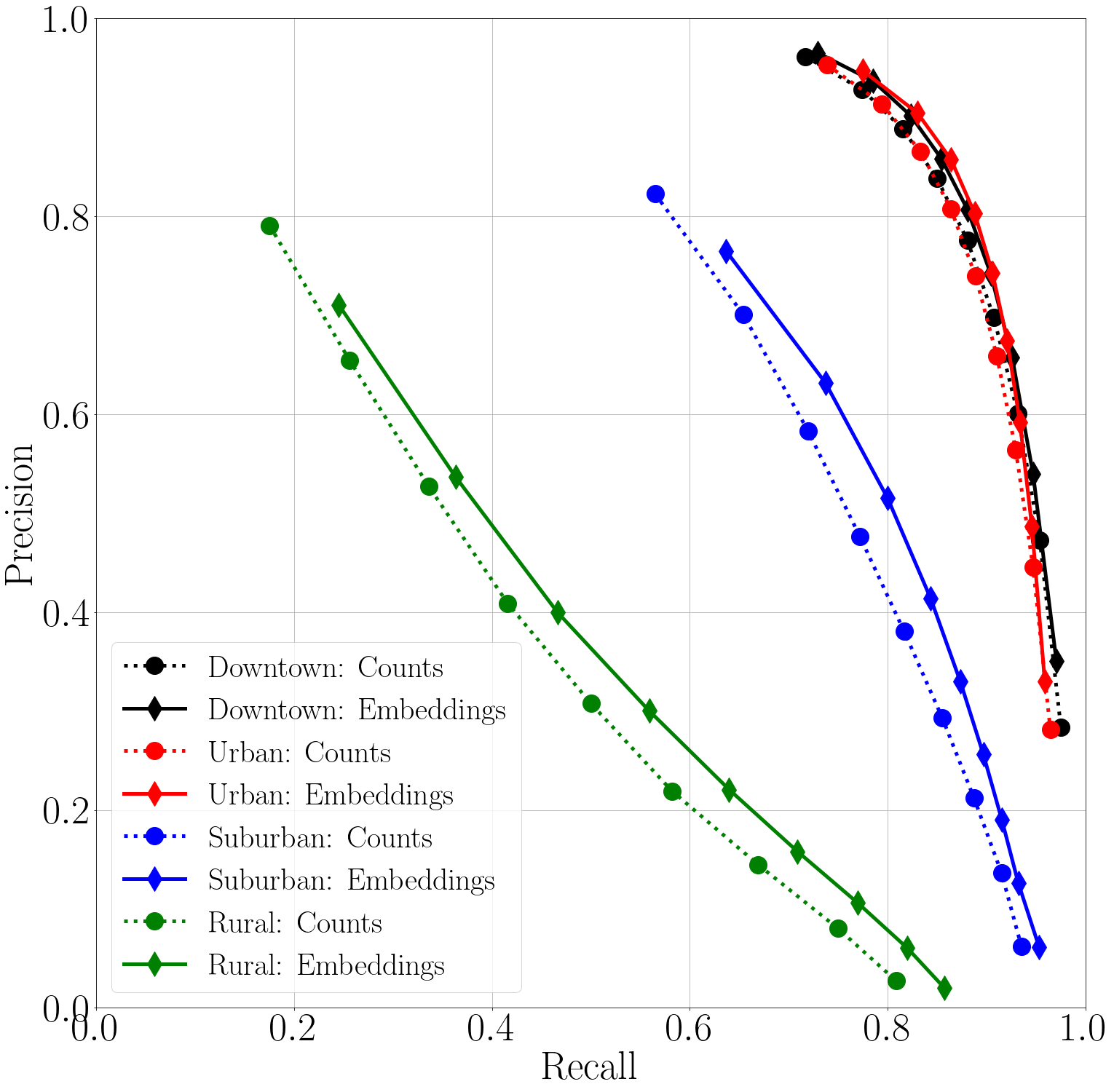}
        \caption{Caption for image}
        \label{fig:pr_curves}
    \end{minipage}
    \begin{minipage}{0.24\textwidth}
        \centering
        \includegraphics[width=0.45\textwidth]{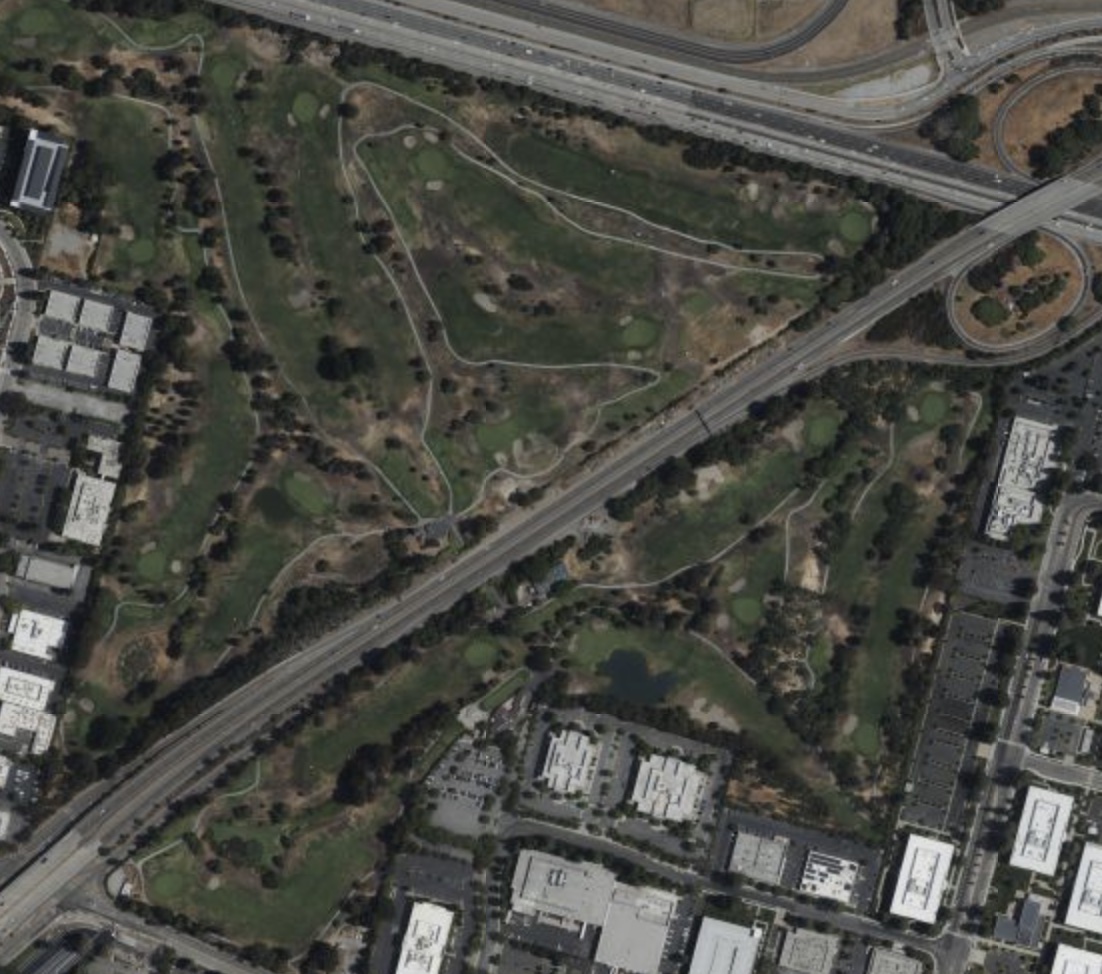}
        \includegraphics[width=0.45\textwidth]{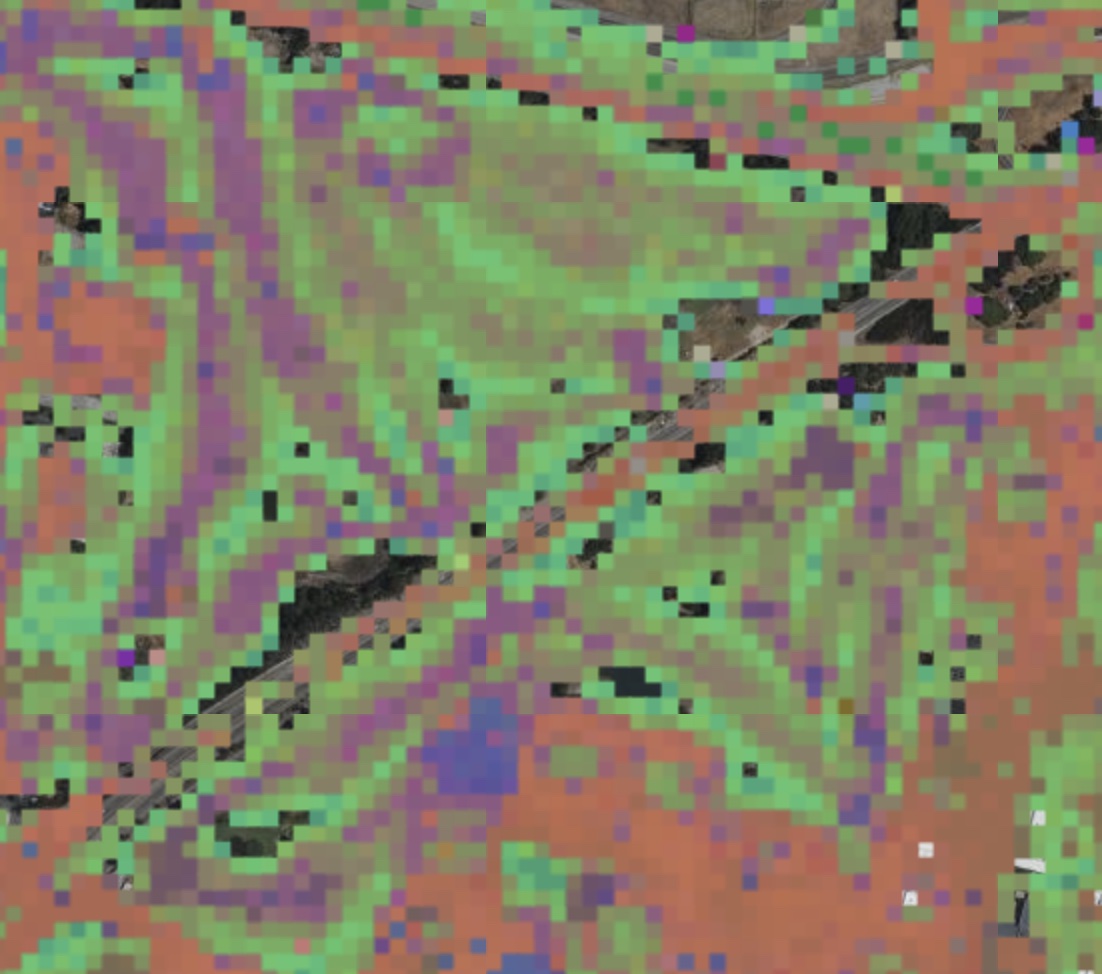}
        
        \includegraphics[width=0.45\textwidth]{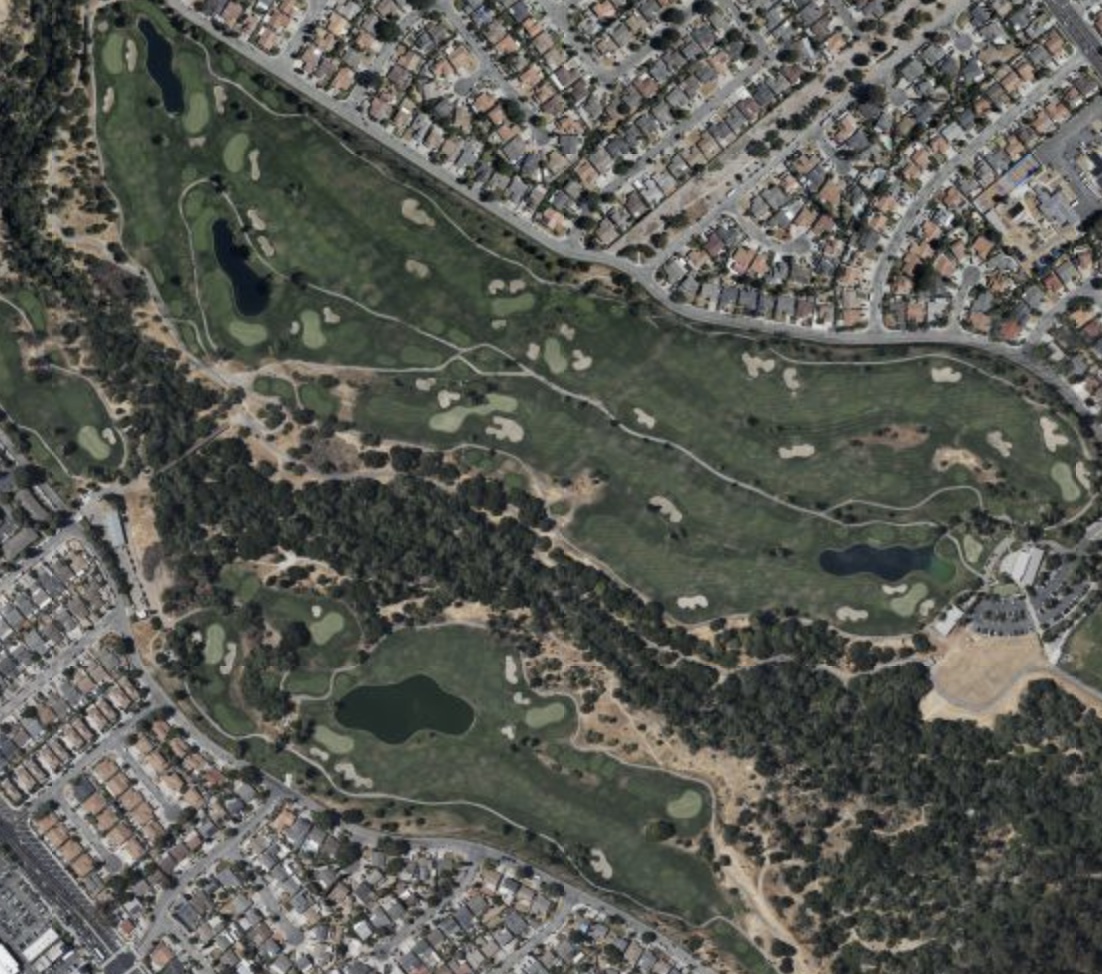}
        \includegraphics[width=0.45\textwidth]{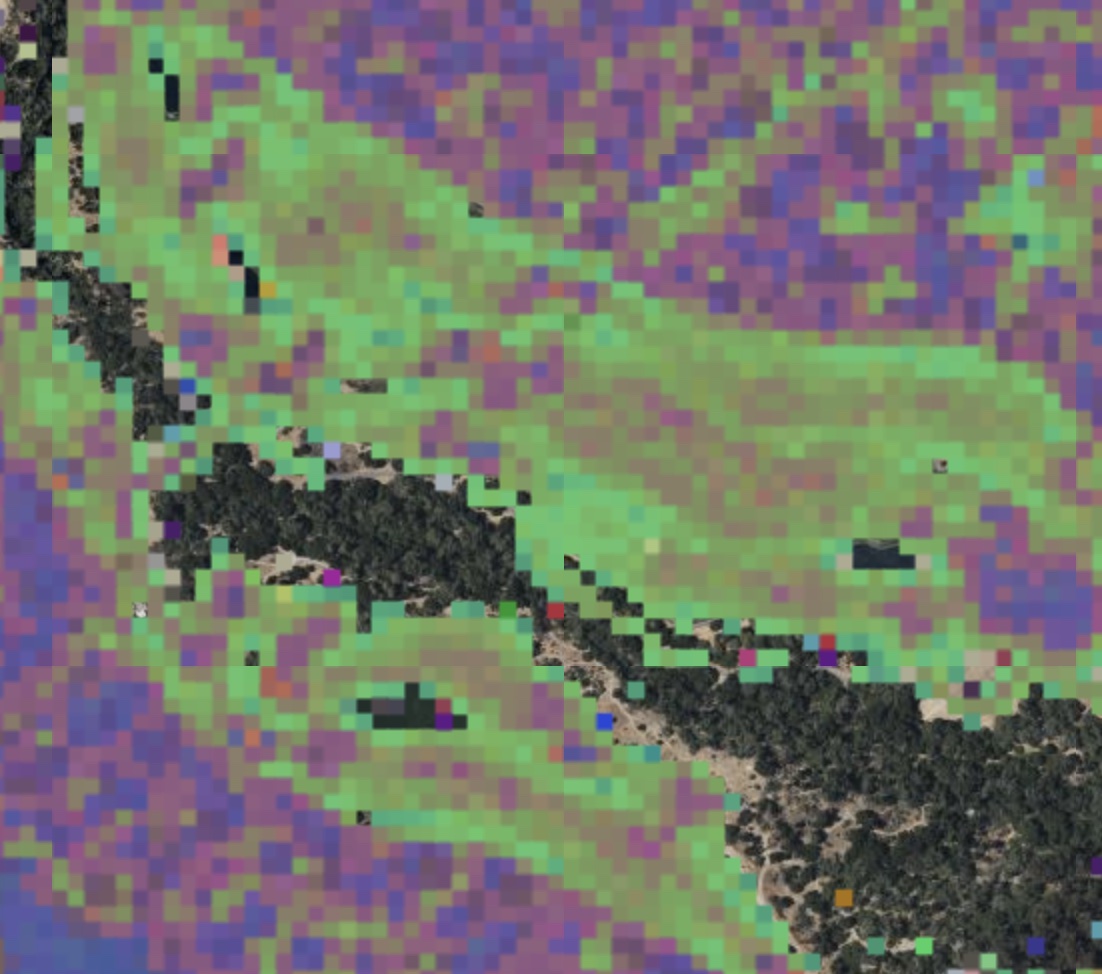}
        \caption{Golf Courses}
        \label{figure:golf}
    \end{minipage}
    \begin{minipage}{0.24\textwidth}
        \centering
        \includegraphics[width=0.45\textwidth]{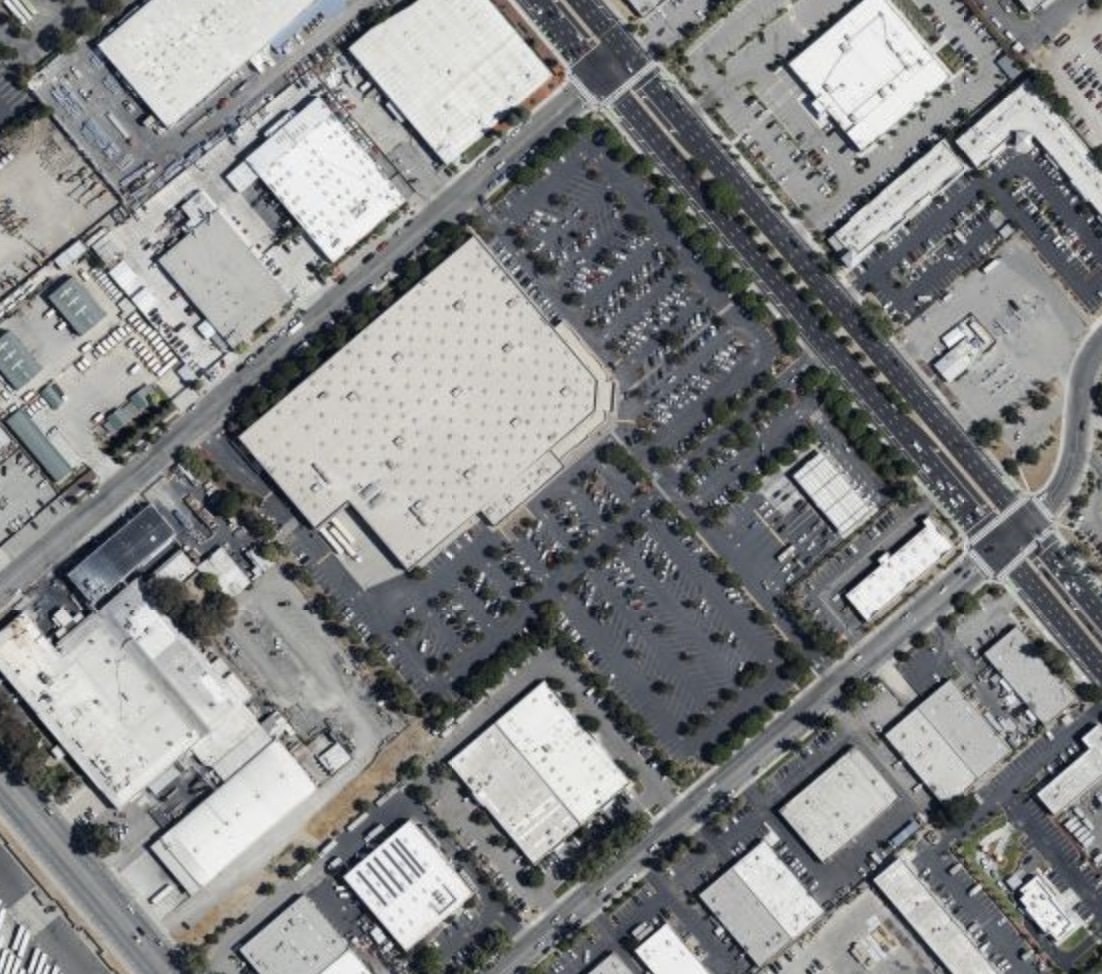}
        \includegraphics[width=0.45\textwidth]{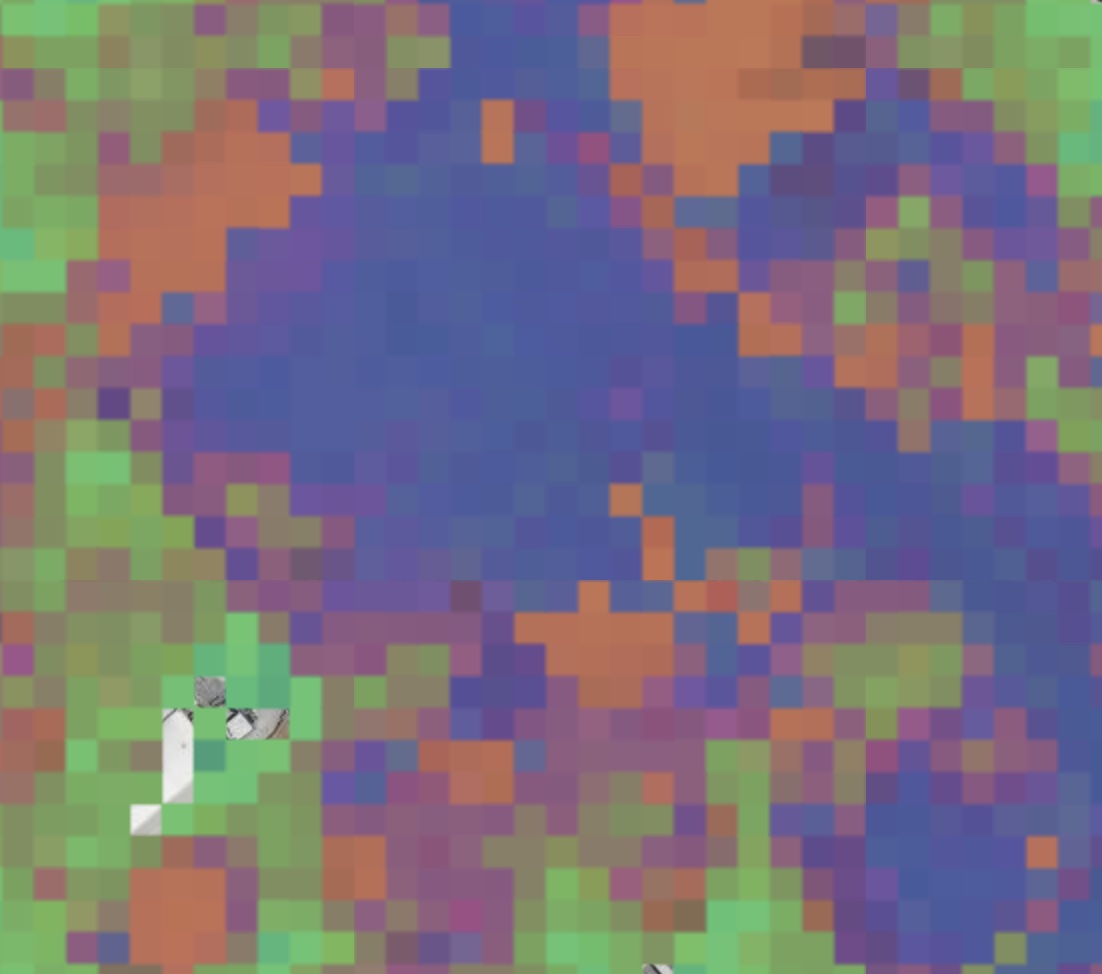}
        
        \includegraphics[width=0.45\textwidth]{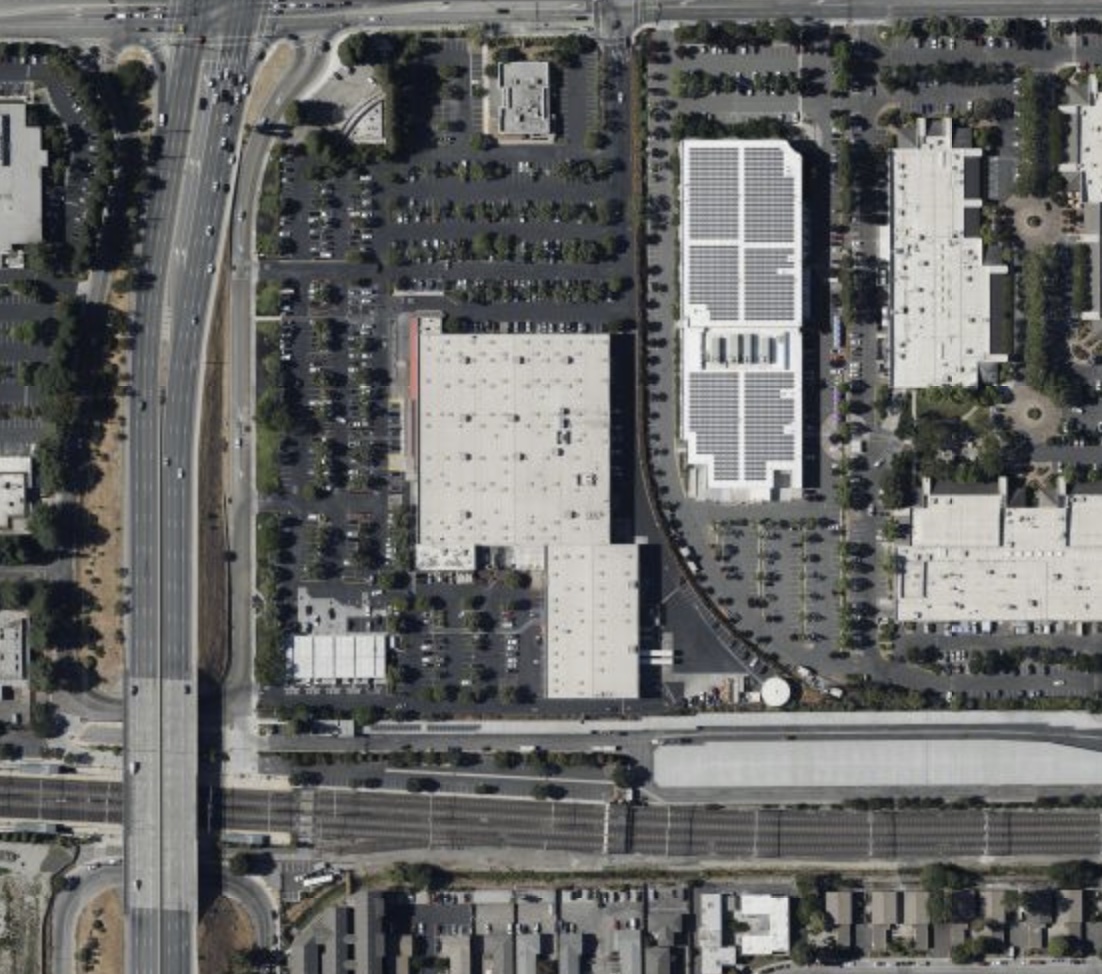}
        \includegraphics[width=0.45\textwidth]{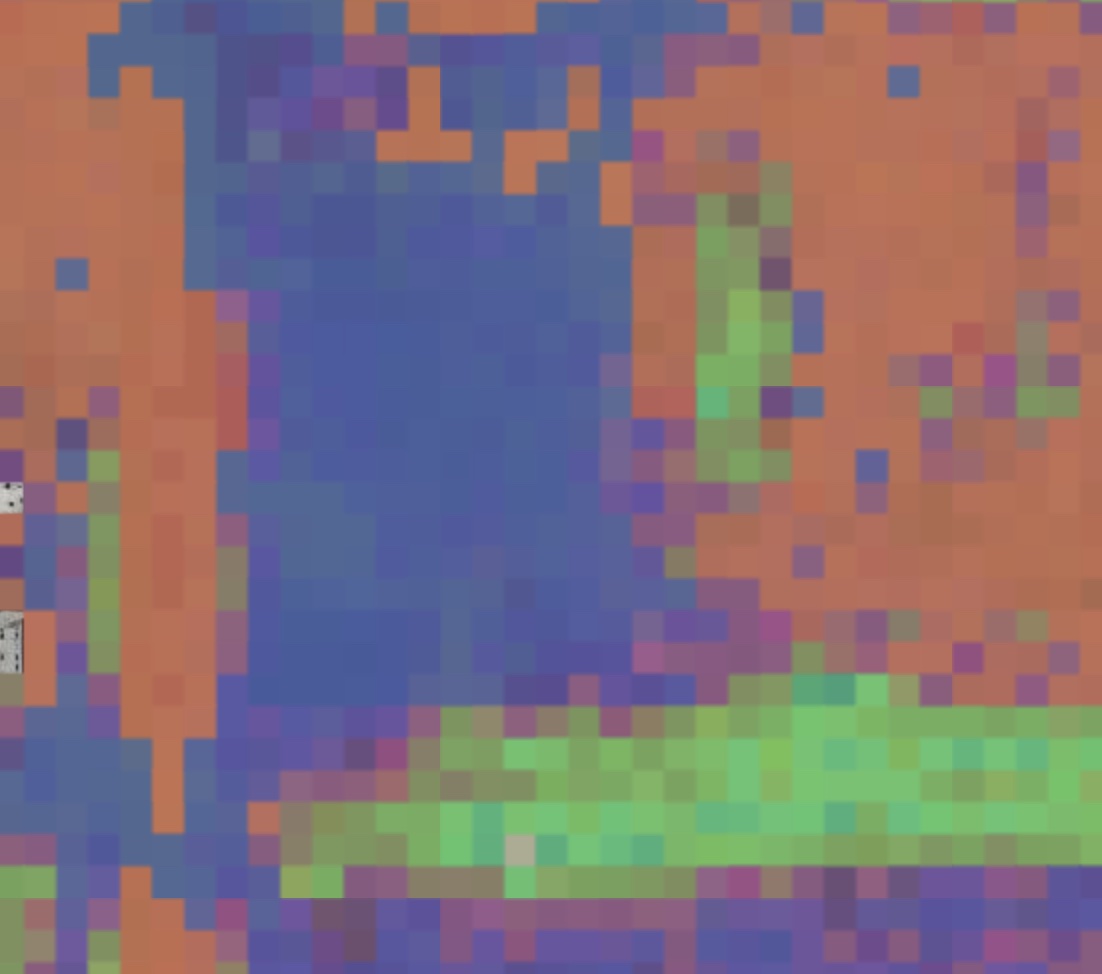}
        \caption{Grocery Shops}
        \label{figure:grocery}
    \end{minipage}
    \begin{minipage}{0.24\textwidth}
        \centering
        \includegraphics[width=0.45\textwidth]{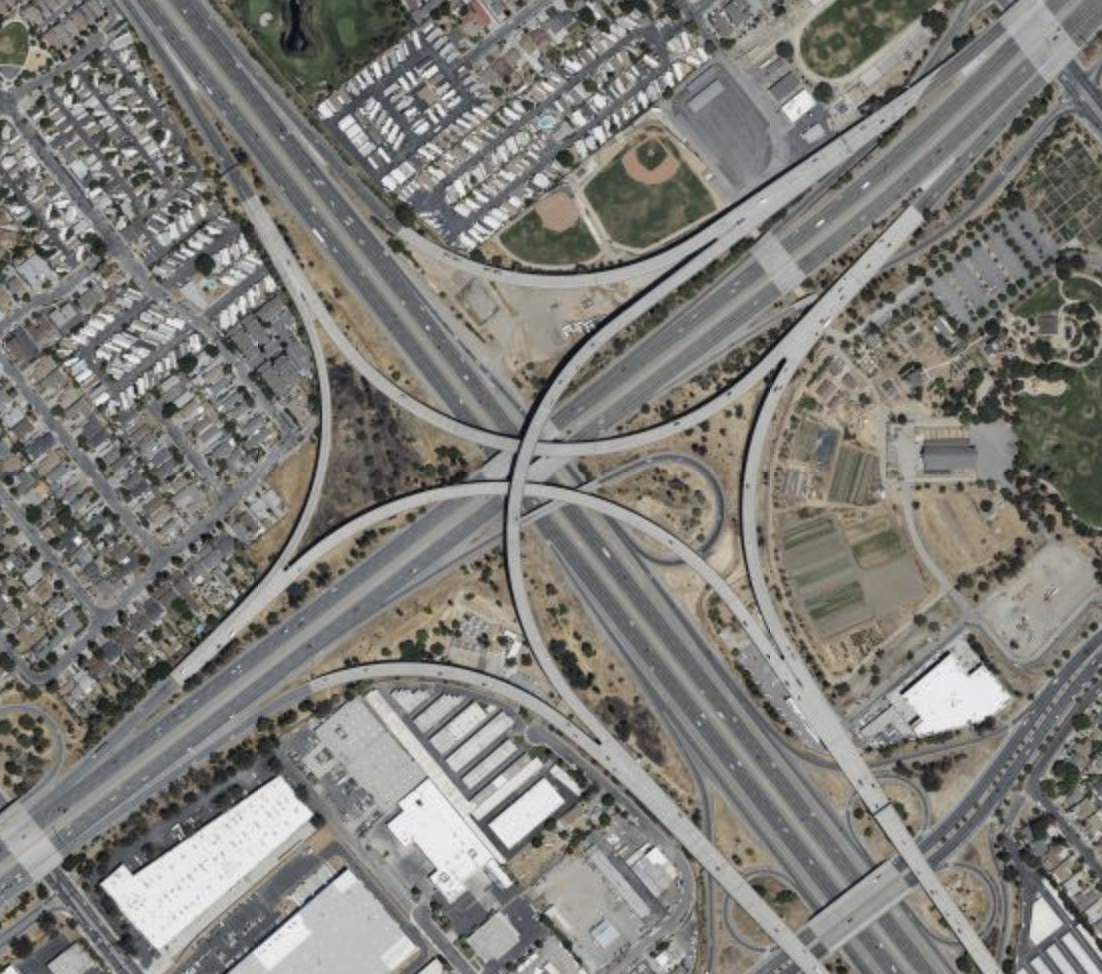}
        \includegraphics[width=0.45\textwidth]{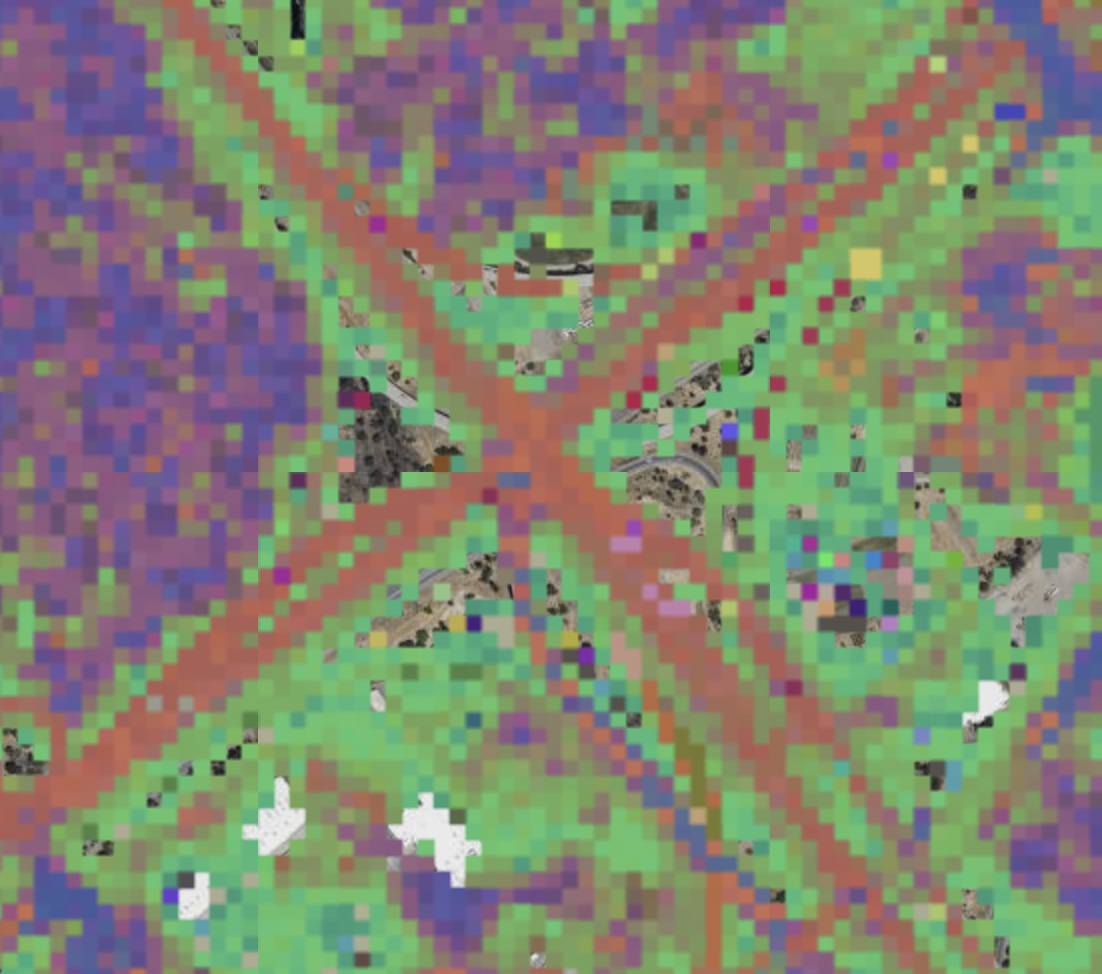}
        
        \includegraphics[width=0.45\textwidth]{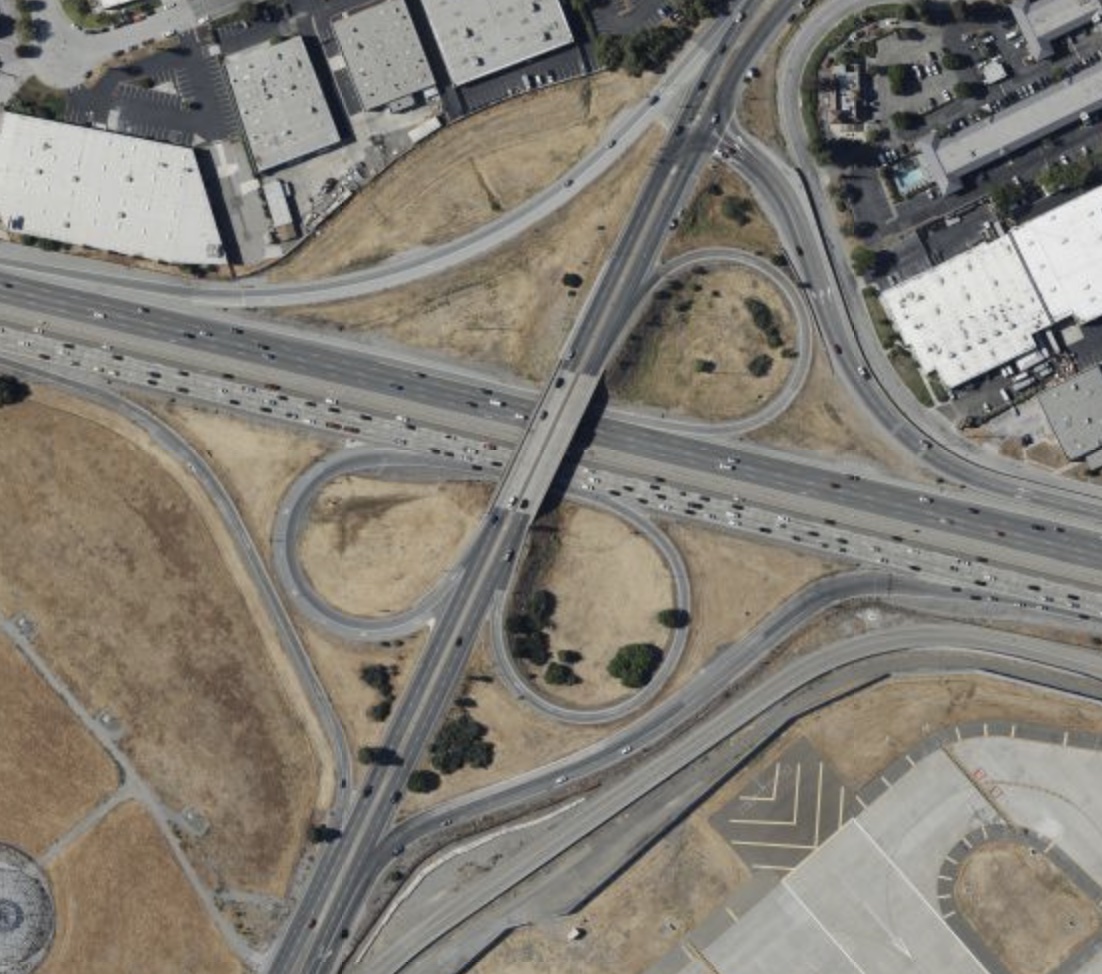}
        \includegraphics[width=0.45\textwidth]{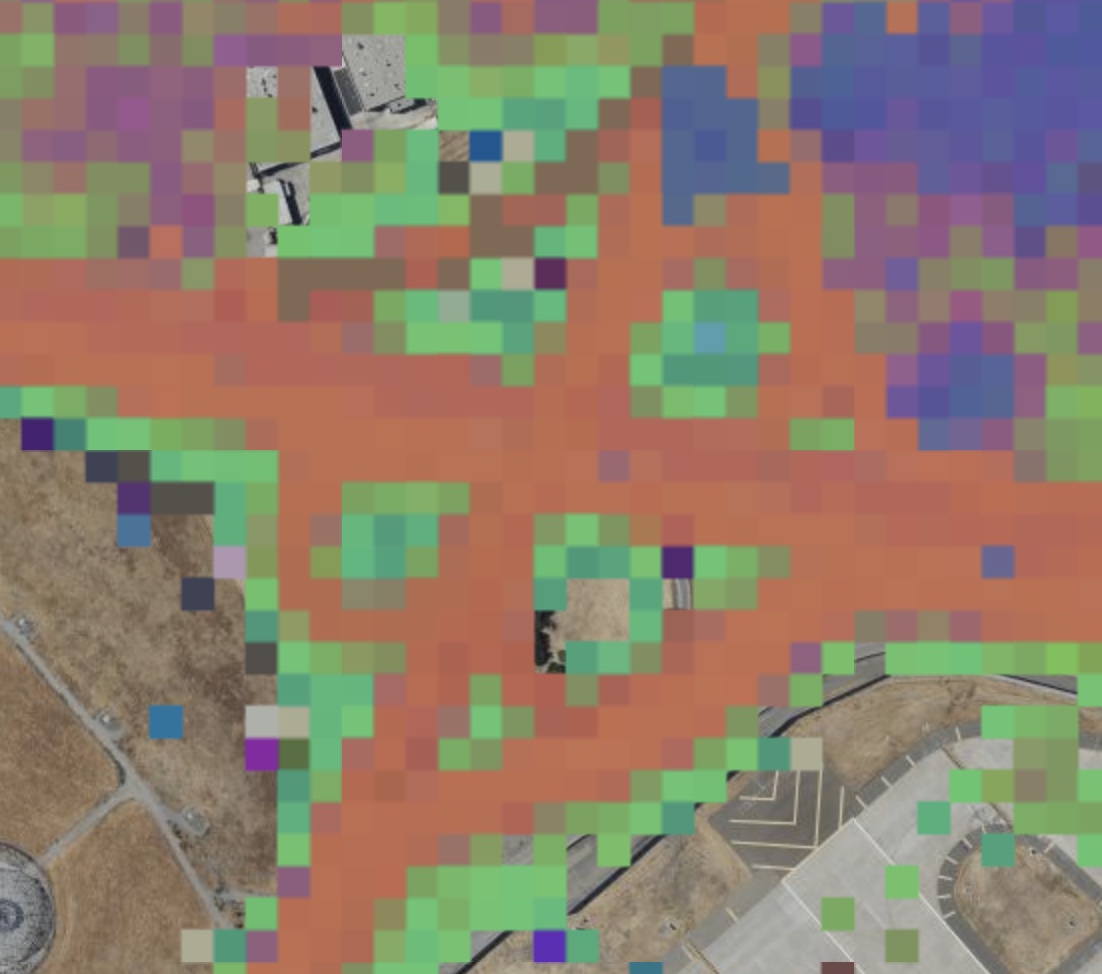}
        \caption{Road Intersections}
        \label{figure:roads}
    \end{minipage}
\end{figure}

\vspace{-0.3cm}
As a verification exercise, we generate 3-dimensional UMAP embeddings from temporal embeddings of locations covering multiple diverse geographies and map the entries to RGB space such that we can color-code temporal patterns on a map, as shown in \cref{umap}. Places with similar temporal patterns are well characterized by the embeddings, which proves that the embeddings capture the spatiotemporal features in the mobility data very well. Through human evaluation, it is found that the color coding strongly correlates to to downtown (red), urban (yellow), suburban (green), and rural (blue) areas. Furthermore, when tested on a held-out test set with labels for the four areas, we see in \cref{fig:pr_curves} that the precision-recall curves of the embedding based classification has higher AUC than using a baseline with DFT or simple activity counts. We use the 16-dimensional embeddings to train a segmentation model and classify landscape into two root categories: residential vs. commercial as shown in \cref{figure::residential_commercial}. Experiments show that the classifier is able to achieve around 85\% precision and recall in urban area, which is higher than using activity density as signal when trained with the same segmentation model. The performance is even higher in suburban and rural areas where activity signal is sparse, proving that temporal patterns buried in the mobility data are a more robust signal than activity density. Temporal embeddings can also be used to classify areas showing similar patterns such as golf courses \cref{figure:golf}, grocery shops \cref{figure:grocery}, road intersections \cref{figure:roads}, among many other categories. These observations conclusively demonstrate that temporal embeddings are more informative, denser representations of temporal patterns in spatiotemporal motion data. Temporal embeddings can also be used to compute semantically meaningful representations of temporal patterns in spatiotemporal data in geographical areas with less traffic or to build computer vision-based models for low-resource geospatial tasks. Additional results of using temporal embeddings in multimodal settings (combining with satellite imagery and road network graph via early fusion) and qualitative comparison of model predictions using temporal embeddings as inputs is presented in \cite{cao2023self} and will be shown in the poster.

%%%%%%%%%%%%%%%%%%%%%%%%%%%%%%%%%%%%%%%%%%%%%%%%%%%%%%%%%%%%%%%%%%%%%%%%
% Section: Bibliography
%%%%%%%%%%%%%%%%%%%%%%%%%%%%%%%%%%%%%%%%%%%%%%%%%%%%%%%%%%%%%%%%%%%%%%%%
\clearpage
\bibliographystyle{abbrv}
\bibliography{Paper_BayLearn2023}

%%%%%%%%%%%%%%%%%%%%%%%%%%%%%%%%%%%%%%%%%%%%%%%%%%%%%%%%%%%%%%%%%%%%%%%%
% End of document
%%%%%%%%%%%%%%%%%%%%%%%%%%%%%%%%%%%%%%%%%%%%%%%%%%%%%%%%%%%%%%%%%%%%%%%%
% End the document
\end{document}